\documentclass[10pt,twocolumn,letterpaper,table,dvipsnames]{article}
\pdfoutput=1 % This tells arXiv to use pdflatex.

\usepackage[utf8]{inputenc}
\usepackage[T1]{fontenc}
\usepackage{textcomp}

\usepackage{3dv}
\usepackage{times}
\usepackage{graphicx}
\usepackage{amsmath}
\usepackage{amssymb}

\usepackage{ifthen}
\newboolean{includeSuppMat}
\setboolean{includeSuppMat}{true} %% <<< use true for arXiv, false for official 3DV version
\newcommand{\ifIncludeSuppMat}[2]{\ifthenelse{\boolean{includeSuppMat}}{#1}{#2}}
\ifIncludeSuppMat{
}{
\usepackage{xr}
\externaldocument[supp-]{WideBaselineSceneFlow-supplemental-3DV2016}
}

\usepackage{microtype}
\usepackage{enumitem}
\usepackage{tikz,pgfplots}
\pgfplotsset{compat=1.3}
\usetikzlibrary{backgrounds,external,calc,arrows,positioning,pgfplots.groupplots}

\tikzset{annotate/.style={ % Leszek's code for annotating images in resolution-independent way.
  execute at begin scope={
    \pgfkeys{src/.store in=\src,src/.value required}
    \pgfkeys{pxwidth/.store in=\pxwidth,pxwidth/.value required}
    \pgfkeys{#1}
    \pgftransformshift{\pgfpointanchor{\src}{north west}}
    \pgfsetyvec{\pgfpointscale{1/\pxwidth}{\pgfpointdiff{\pgfpointanchor{\src}{north west}}{\pgfpointanchor{\src}{south west}}}}
    \pgfsetxvec{\pgfpointscale{1/\pxwidth}{\pgfpointdiff{\pgfpointanchor{\src}{north west}}{\pgfpointanchor{\src}{north east}}}}
}}}
\makeatletter\tikzset{add font/.code={\expandafter\def\expandafter\tikz@textfont\expandafter{\tikz@textfont#1}}}\makeatother
\tikzset{every node/.style={add font=\sffamily}}
\tikzset{every picture/.style={tight background,>=stealth}}

\usepackage[pagebackref,breaklinks,colorlinks,bookmarks=false]{hyperref}

\usepackage[numbers,sort]{natbib}
\usepackage[normalem]{ulem}
\usepackage[capitalise]{cleveref}
\Crefname{figure}{Figure}{Figures}
\crefformat{equation}{Equation~#2#1#3}
\crefmultiformat{equation}{Equations~#2#1#3}{ and~#2#1#3}{, #2#1#3}{ and~#2#1#3}

\newcommand*\dif{\mathop{}\!\mathrm{d}} %% from <http://tex.stackexchange.com/a/95681/>

\newcommand{\inlineheading}[1]{\vspace{0.5em}\noindent\textbf{#1\hspace{0.5em}}}

\threedvfinalcopy % *** Uncomment this line for the final submission

\def\threedvPaperID{71} % *** Enter the 3DV Paper ID here

\ifthreedvfinal\fi
\begin{document}

%%%%%%%%% TITLE
\title{Dense Wide-Baseline Scene Flow From Two Handheld Video Cameras}

\author{%
Christian Richardt $^\text{1, 2, 3}$\qquad%\\
Hyeongwoo Kim $^\text{1}$\qquad%\\
Levi Valgaerts $^\text{1}$\qquad%\\
Christian Theobalt $^\text{1}$%
%\\\vspace*{1pt}\\% Matches the spacing in review version.
\\[0.5em]
$^\text{1}$ Max Planck Institute for Informatics\quad%
$^\text{2}$ Intel Visual Computing Institute\quad%
$^\text{3}$ University of Bath%
}

\hypersetup{
	pdftitle={Dense Wide-Baseline Scene Flow From Two Handheld Video Cameras},
	pdfauthor={Christian Richardt, Hyeongwoo Kim, Levi Valgaerts, Christian Theobalt}
}

\maketitle

\begin{abstract}
We propose a new technique for computing dense scene flow from two handheld videos with wide camera baselines and different photometric properties due to different sensors or camera settings like exposure and white balance.
Our technique innovates in two ways over existing methods:
(1) it supports independently moving cameras, and
(2) it computes dense scene flow for wide-baseline scenarios.
We achieve this by combining state-of-the-art wide-baseline correspondence finding with a variational scene flow formulation.
First, we compute dense, wide-baseline correspondences using \textsc{daisy} descriptors for matching between cameras and over time.
We then detect and replace occluded pixels in the correspondence fields using a novel edge-preserving Laplacian correspondence completion technique.
We finally refine the computed correspondence fields in a variational scene flow formulation.
We show dense scene flow results computed from challenging datasets with independently moving, handheld cameras of varying camera settings.
\end{abstract}

%%==================================================================================================
\section{Introduction}
\label{sec:intro}

%% The problem
%% (2) general motivation for the problem
A variety of methods to reconstruct space-time coherent geometry of dynamic scenes from multiple video or depth cameras have been proposed, such as performance capture methods in vision and graphics \cite{MagnoGST2015}.
Many video-based methods rely on a combination of correspondence finding between camera views to capture shape, and temporal correspondence finding to establish temporal coherence in reconstructions.
A deforming template model is often used to assist shape and motion estimation.
Most existing approaches are limited to controlled indoor environments with rather dense static camera setups and controlled backgrounds, or are limited to reconstruction of a few dynamic foreground objects, and thus cannot capture the entire scene \cite{MustaKGH2015}.
In recent years, the trend for dynamic scene reconstruction has been towards increasingly unconstrained capturing of scenes from multiple videos.
For instance, first template-based methods to reconstruct humans or faces from stereo cameras with a fixed baseline and in less controlled surroundings, including outdoor scenes, were proposed \cite{ValgaWBST2012,WuSVT2013}.

\noindent
The widespread proliferation of mobile video cameras, particularly in mobile phones, has accelerated this trend and led to an explosion of recorded video content, which could be used for dynamic scene reconstruction and free-viewpoint rendering, for instance of music or sports events filmed by several spectators \cite{BallaBPP2010,LipskKM2014}.
However, videos recorded with mobile cameras pose a new, generalised wide-baseline stereo problem \cite{MishkMPL2015}.
Correspondences have to be robust to both independent camera motion with wide geometric baselines, as well as starkly differing image characteristics due to different sensors and camera settings, such as exposure and white balance (aka the `\textsc{WgsBS}' problem \cite{MishkMPL2015}).
Most existing techniques assume static camera setups, and only few handle moving stereo rigs with fixed baselines, even though videos captured with independently moving handheld cameras of potentially different type are now the norm in practice.
In addition to handling these independent cameras, approaches shall also be able to reconstruct dense geometry in both space and time, i.e. shape and motion of all objects in the scene, not only foreground objects for which a template needs to be created in a complex pre-processing step.
They also need to succeed with wide-baseline recordings and footage recorded under other adverse effects such as changing lighting and frequent occlusions.
Empowering dynamic scene reconstruction methods to handle such scenes requires algorithmic innovation on several ends.
In this paper, we take a step towards this goal by proposing a new solution to dense correspondence finding in more general settings.

%% Limitations of existing solutions.
%% (3) what are the alternative approaches to this problem 
Most dynamic scene reconstructions are based on the estimation of the 3D scene motion over time, which is known as \emph{scene flow}, a term coined by Vedula et al. \cite{VedulBRCK2005} in analogy to the term `optical flow' for the motion over time.
Our goal is to compute dense scene flow of general dynamic scenes from two handheld videos of independently moving cameras with wide baseline in terms of both camera geometry and sensor characteristics.
This is not supported by current scene flow techniques:
dense approaches generally rely on narrow camera baselines \cite{HugueD2007,ValgaBZWST2010,BashaMK2013}, and wide-baseline approaches are not dense as they use sparse scene representations such as voxels or particles \cite{VedulBRCK2005,DeverMG2006,HadfiB2014}.
In addition, previous techniques do not support different sensors and handheld videos, as they assume constant camera calibration over time.

%% Key idea / our approach.
%% (1) high-level description of approach
%% (4) what is different about your approach
%% (5) what is your silver bullet or a key idea/concept
\noindent
We propose a technique that overcomes these limitations by combining wide-baseline correspondence finding with a dense, variational scene flow computation approach that jointly estimates dense correspondence fields across camera views and within camera views of two subsequent time steps, even if sensor or image characteristics differ notably between cameras.
%
%Our approach can also implicitly estimate camera geometry alongside dense correspondences.
%
Our technical contributions are
(1) a novel correspondence finding technique that uses \textsc{daisy} descriptors \cite{TolaLF2010} for wide-baseline matching in both space (between cameras) and time (in the same camera view), and is optimised using PatchMatch belief propagation (\textsc{pmbp}) \cite{BesseRFK2014}, and
(2) an edge-preserving Laplacian correspondence completion technique.
We show dense scene flow results, alongside with dense stereo geometry, computed from challenging independently moving handheld video datasets with medium to wide camera baselines, which we will make publicly available.

%%==================================================================================================
\section{Related work}
\label{sec:related}

Scene flow describes the motion within a scene over time, specifically the motion of every visible 3D point between two time steps.
Many techniques have been proposed to compute the scene flow from two or more videos, including
voxel coloring from dense in-studio camera setups \cite{VedulBRCK2005}, surfel tracking \cite{DeverMG2006}, and growing correspondence seeds \cite{CechSH2011}.
Scene flow was also computed as part of non-rigid scene registration \cite{BashaAHM2012}, and by means of particle-based estimation~\cite{HadfiB2014}.
However, the most common class of scene flow approaches are variational methods \cite{HugueD2007,PonsKF2007,BashaMK2013,ValgaBZWST2010,WedelBVRFC2011,FerstRRB2014,SunSP2015,JaimeSSGC2015,ThiesZRTG2016}, which provide dense, continuous and strongly regularised solutions.
Some techniques enforce motion priors such as affine \cite{ZhangK2003} or piece-wise rigid motions \cite{VogelSR2015,MenzeG2015,JaimeSSGC2015}, but these are violated by the non-rigid scenes we are targeting with our approach.
Many recent techniques also build on RGB-D data obtained from consumer depth sensors \cite{LetouPB2011,HerbsRF2013,HadfiB2014,HornaFR2014,QuiroBDC2014,FerstRRB2014,SunSP2015,JaimeSSGC2015,ZanfiS2015}.
However, these approaches are limited to indoor use due to the depth sensors, while we target general, unconstrained outdoor settings with normal video cameras.
Like most techniques, we compute scene flow between exactly two time steps; only few techniques enforce temporal consistency over multiple time steps \cite{HungXJ2013,VogelRS2014}. % GarriVWT2013
We also assume synchronised input videos, which can be achieved in hardware or software \cite{MeyerSMP2008,HasleRTWGS2009,ElhaySKST2012,GaspaOF2014}.

Even though some dense variational approaches are able to handle moving cameras \cite{ValgaBZWST2010,WedelBVRFC2011}, they assume a static camera rig with a fairly narrow baseline (10–30\,cm) and fail with wider camera baselines or when cameras are moving independently, as in our case.
On the other hand, methods that succeed on wider baselines only reconstruct sparse correspondences.
In contrast, our method captures dense scene flow and stereo geometry also with wider baselines and independently moving cameras.

Scene flow estimation is also related to non-rigid structure from motion \cite{AvidaS2000,HartlV2008,GargRA2013b,ParkSMS2010,RusseYA2014,DaiLH2014,YuRCA2015,ZhengJDF2015}, % VidalA2006,AkhteSKK2011,AgudoMAC2014
but these approaches make strong prior assumptions about scene motion models and work best with small displacements between video frames.
Our work is also related to spatio-temporal stereo matching, which has been demonstrated for static camera setups and controlled scenes \cite{ZhangCS2003,RichaODCD2010,JiangLTZB2012}. % SizinW2012,ShinY2015
As stated in the introduction, scene flow is an important ingredient for many applications, including 3D motion understanding \cite{WedelBVRFC2011,MenzeG2015}, facial performance capture \cite{ValgaWBST2012,WuSVT2013} % GarriVWT2013
and free-viewpoint video \cite{LipskKM2014}. % LipskLBSM2010
This paper paves the way to lifting these applications to the case of independent handheld video in general scenes.

Wide-baseline matching addresses the difficult task of finding corresponding points in potentially very different viewpoints.
The most robust matching results have been achieved using affinely invariant features \cite{MoreeP2007,TuyteM2008}.
While robust, these techniques are very sparse as they only produce a few hundred correspondences per image pair.
Follow-up work hence explored densification using a multi-resolution variational formulation \cite{StrecTV2003} or match propagation \cite{KannaB2007}.
To avoid a separate densification step, the \textsc{daisy} descriptor \cite{TolaLF2010} we use was specifically designed for dense wide-base\-line matching \cite{TolaSF2012}.
We thus use insights from the design of \textsc{daisy} and adapt it to the case of dense matching across camera views \textit{and} over time.

%%==================================================================================================
% !TEX root = WideBaselineSceneFlow-3DV2016.tex
% !TeX spellcheck = en_GB
% !TEX spellcheck = en_GB
%%==================================================================================================
\section{Method}
\label{sec:method}

Our approach computes a dense reconstruction of geometry and scene flow from dynamic scenes casually captured with two handheld video cameras, without imposing any specific assumptions about the scene structure or camera motion.
Cameras can differ in make and sensor characteristics, and our method is one of the first to tolerate notable appearance differences between videos.

We distinguish three kinds of correspondences, which we all call \emph{flows} for simplicity: \emph{stereo flow} is the correspondence between images from different cameras at the same time, \emph{optical flow} is correspondence over time within a camera, and \emph{scene flow} describes the 3D motion over time.
We use a pipeline with four main stages:
\begin{enumerate}[itemindent=1em,parsep=-0.2em]
	\item synchronisation and calibration,
	\item correspondence finding,
	\item occlusion filling, and
	\item scene flow computation.
\end{enumerate}
We first synchronise the input videos and calibrate the cameras, before we estimate bidirectional correspondences between pairs of images using the same novel technique for both stereo flow (between cameras at the same time) and optical flow (same camera over time).
The stereo correspondences are then postprocessed by invalidating and filling occlusions using a novel edge-preserving scheme based on local linear regression.
We finally refine the computed correspondence fields in a variational scene flow formulation.

\inlineheading{Calibration}
We assume that our input videos have known, fixed camera intrinsics, and are synchronised temporally, which can be done automatically by video-based methods \cite{ElhaySKST2012,GaspaOF2014}.
Similar to most approaches looking into wide-base\-line stereo reconstruction, we calibrate the moving cameras extrinsically by undistorting all input video frames and then estimating the camera geometry using structure-from-motion techniques with fixed intrinsics \cite{Wu2011,ZouT2013}.

%%--------------------------------------------------------------------------------------------------
\subsection{Correspondence finding with \textbf{\textsc{daisy + pmbp}}}
\label{sec:DaisyPMBP}

\begin{figure}
	\input{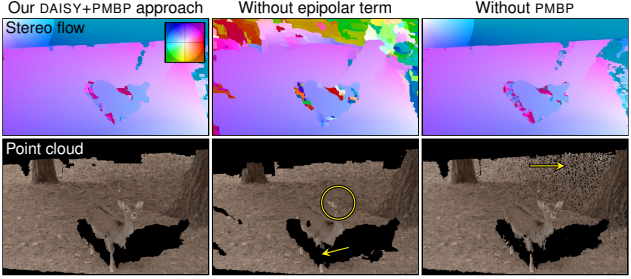}
	\caption{\label{fig:DaisyPMBP-ablation}%
		Our \textsc{daisy+pmbp} correspondences compared to leaving out the epipolar term ($w_\text{E} \!=\! 0$) or \textsc{pmbp} ($w_\text{p} \!=\! 0$), which results in missing parts like the deer's head (centre) and incoherent geometry (right), respectively.
	}\vspace{-1em}
\end{figure}

%% Matching cost.
Our correspondence finding strategy is based on the \textsc{daisy} descriptor by Tola et al. \cite{TolaLF2010}.
In contrast to most other descriptors, such as \textsc{sift} or \textsc{surf}, which were designed for describing sparse interest points \cite{MoreeP2007,TuyteM2008}, \textsc{daisy} was designed for finding dense correspondences, specifically in wide-baseline scenarios.
The \textsc{daisy} descriptor encodes local appearance using image gradient histograms across different gradient orientations.
The gradients are computed at different image scales for different points around the descriptor location, depending on their distance to it.
This results in a flower-like arrangement – hence the name \textsc{daisy}.
However, pixel-wise local matching can give rise to spatial inconsistencies.
We propose a matching scheme that establishes geometric and spatial coherence by introducing an epipolar energy term and a global matching scheme (see \cref{fig:DaisyPMBP-ablation}).

We use \textsc{daisy} as the key ingredient in our matching cost $c\! \left( \mathbf{x}, \mathbf{y} \right)$ between image locations $\mathbf{x}$ and $\mathbf{y}$ in different images:
\begin{equation}
\label{eq:MatchingCost}
c\! \left( \mathbf{x}, \mathbf{y} \right) =
c_\text{D}\! \left( \mathbf{x}, \mathbf{y} \right) +
c_\text{C}\! \left( \mathbf{x}, \mathbf{y} \right) +
c_\text{E}\! \left( \mathbf{x}, \mathbf{y} \right) \! \text{,}
\end{equation}
which combines the \textsc{daisy} descriptor difference $c_\text{D}$ with a colour consistency term $c_\text{C}$, and an epipolar term $c_\text{E}$ (only used for stereo flows).
We next describe each of these terms, and then discuss how we minimise the matching costs across all pixels to compute correspondence fields.

%% Daisy term.
The \textsc{daisy} term measures the dissimilarity of local image regions using the difference between the \textsc{daisy} descriptors $\mathbf{D}(I, \mathbf{x})$ computed at the two considered locations $\mathbf{x}$ and $\mathbf{y}$ in images $I_1$ and $I_2$:
\begin{equation}
\label{eq:DaisyTerm}
c_\text{D}\! \left( \mathbf{x}, \mathbf{y} \right) = w_\text{D} \cdot \left\| \mathbf{D}(I_1, \mathbf{x}) - \mathbf{D}(I_2, \mathbf{y}) \right\|_2^2 \text{,}
\end{equation}
where $w_\text{D}$ is the weight for this term.
As suggested by the authors of \textsc{daisy}, we orient descriptors along the epipolar lines when epipolar geometry is given.
For example, the descriptor at $\mathbf{x}$ is oriented along the epipolar line $\mathbf{l} \!=\! \mathbf{F}^\top \mathbf{y}$.
We use the authors' implementation of the descriptor (aka `libdaisy'), but to preserve more geometric details, we use a smaller footprint of only two rings with a radius of 10 pixels for stereo flow computation.

\begin{figure}
	\input{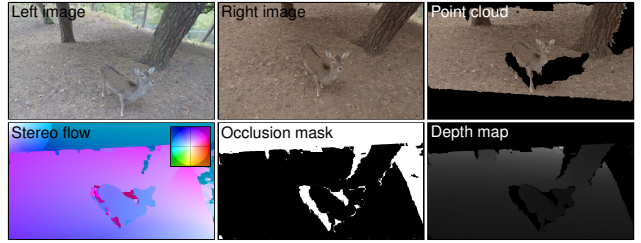}
	\caption{\label{fig:DaisyPMBP}%
		Two input frames from our handheld \textsc{deer} dataset with different exposure and white balance, which are captured about 65\,cm apart with an angle of about 13 degrees between the optical axes, and the resulting triangulated point cloud (top right).
		Below: estimated stereo correspondences, occlusion mask and depth map.
%		This sort of deer (dama dama) should have a shoulder height of 80-100cm (measured as 1.6 units in Meshlab).
%		0131. F: GOPRO_B-medium => GOPRO_C-medium
%		- baseline:   1.142 units ~=  57-71 cm
%		- camB-nose:  7.832 units ~= 392-490 cm
%		- camC-nose:  8.015 units ~= 401-501 cm
%		- cam  angle:  10.01 deg
%		- nose angle:   8.16 deg
	}\vspace{-1em}
\end{figure}

%% Colour term.
The colour consistency term $c_\text{C}$ helps to disambiguate image regions with similar gradient distributions, for example in areas of constant colour.
In those regions, \textsc{daisy} descriptors are often similar, but the colour term penalises different colours that would result in mismatches.
We compute the colour term using
\begin{equation}
\label{eq:ColourTerm}
c_\text{C}\! \left( \mathbf{x}, \mathbf{y} \right) = w_\text{C} \cdot \left\| \mathbf{A} I_1( \mathbf{x}) + \mathbf{a} - I_2(\mathbf{y}) \right\|_2 \text{,}
\end{equation}
where $w_\text{C}$ is the colour term weight, and the 3$\times$3 matrix $\mathbf{A}$ and offset $\mathbf{a}$ apply an affine colour transformation that adjusts the colours in image $I_1$ to be closer to those in image $I_2$ (in RGB colour space).
This compensates for differences in exposure and white balance between the images (e.g. see \cref{fig:DaisyPMBP}).
At first, we initialise $\mathbf{A}$ to the identity matrix and $\mathbf{a}$ to the zero vector, but in later iterations, we estimate them with a least-squares fit to the colours of corresponding pixels.

%% Epipolar term.
The epipolar term $c_\text{E}$ measures how well two image points $\mathbf{x}$ and $\mathbf{y}$ satisfy the epipolar geometry defined by the fundamental matrix $\mathbf{F}$.
In stereo correspondence, this term helps to constrain correspondences to lie close to each others' epipolar lines.
This reduces the search space and mismatches.
We use the Sampson distance as described by Hartley and Zisserman \cite[Section 11.4.3]{HartlZ2004}:
\begin{equation}
\label{eq:EpipolarTerm}
c_\text{E}\! \left( \mathbf{x}, \mathbf{y} \right) = \frac{
	w_\text{E} \cdot ( \mathbf{y}^\top \mathbf{F} \mathbf{x} ) ^ 2
}{
	(\mathbf{F} \mathbf{x})_1^2 +
	(\mathbf{F} \mathbf{x})_2^2 +
	(\mathbf{F}^\top \mathbf{y})_1^2 +
	(\mathbf{F}^\top \mathbf{y})_2^2
} \text{,}
\end{equation}
where $w_\text{E}$ is the weight for the epipolar term, and $(\mathbf{F} \mathbf{x})_k^2$ represents the square of the $k$-th entry of the vector $\mathbf{F} \mathbf{x}$.

%% PMBP.
We want to minimise the matching cost in \cref{eq:MatchingCost} across all pixels in a locally smooth way.
For this, we chose a variant of PatchMatch \cite{BarneSGF2010}, as its stochastic initialisation provides good initial correspondences even in challenging wide-baseline cases.
Smoothness of the correspondence field is encouraged using the PatchMatch belief propagation (\textsc{pmbp}) technique by Besse et al. \cite{BesseRFK2014}, which introduces a pairwise term $p$ for regularisation into the energy formulation:
\begin{equation}
\label{eq:PMBP}
E%(\{\mathbf{y}_i\})
= \underbrace{\sum_i c\! \left( \mathbf{x}_i, \mathbf{y}_i \right) \vphantom{\sum_{j \in N(i)}}}_\text{\small unary terms} + 
\underbrace{\sum_i \sum_{j \in N(i)} p\! \left( \mathbf{x}_i, \mathbf{y}_i, \mathbf{x}_j, \mathbf{y}_j \right)}_\text{\small pairwise terms} \text{\!,}
\end{equation}
where $N(i)$ represents the set of 4-neighbours of pixel $i$.
We use the truncated squared difference between flows as our pairwise term to enforce smoothness:
{\small\begin{equation}
\label{eq:PairwiseTerm}
p\! \left( \mathbf{x}_1, \mathbf{y}_1, \mathbf{x}_2, \mathbf{y}_2 \right) \!=\! \min\!\left(\!\tau_\text{p}, w_\text{p} \!\cdot\! \left\|
	(\mathbf{y}_1 \!-\! \mathbf{x}_1) \!-\! (\mathbf{y}_2 \!-\! \mathbf{x}_2)
\right\|_2^2 \right) \!\text{,}
\end{equation}\par}
\noindent
using the threshold $\tau_\text{p}$ and weight $w_\text{p}$.
We use \textsc{pmbp} to compute bidirectional correspondences between each pair of images, so that we can easily check for their consistency in subsequent computation steps.

%%% USAC.
%Without a given extrinsic calibration \cite[e.g. from][]{Wu2011,ZouT2013}, we estimate the fundamental matrix from the computed stereo flow after each run of \textsc{pmbp} with our matching cost.
%%
%To robustly estimate the fundamental matrix, even for scenes with a dominant plane, we use \textsc{usac} \cite{RagurCPMF2013}, a framework for random sample consensus with improvements in terms of accuracy, efficiency and robustness over standard \textsc{ransac} \cite{FischB1981}.
%%
%We first subsample correspondences to a around 30,000 to speed up computation, and then apply \textsc{usac} with an inlier threshold of 1 pixel and a confidence of 99.9\%.
%%
%Finally, we refit the fundamental matrix to all inliers using the 8-point algorithm \cite[Section 11.2]{HartlZ2004}.
%%
%This ensures a robust estimation of the epipolar geometry even in the presence of many mismatches.

%% Iterations and parameters.
We refine our correspondences over multiple passes in which we estimate the affine colour transform $[\mathbf{A}\ \mathbf{a}]$ after each run of \textsc{pmbp}.
We use the same settings for all sequences, which illustrates the stability of our approach across differing camera responses.
We keep the parameters $(w_\text{D}, \tau_\text{p}) \!=$ (1,~50) constant for all passes, but vary the other parameters.
For stereo correspondence, we use four passes, each with two iterations of \textsc{pmbp}.
In the first pass, we use equal weights $w_\text{D} \!=\! w_\text{C} \!=\! w_\text{E} \!=$1, but set the colour weight $w_\text{C}$ to 10 after estimating the colour transform in the first pass.
We also increase the pairwise weight $w_\text{p}$ across passes (using values 0.01, 0.02, 0.1, 1) to more strongly enforce smoothness as matching progresses.
This produces results such as shown in \cref{fig:DaisyPMBP}.
For optical flow, we use only two passes, with 6 and 4 \textsc{pmbp} iterations respectively, with parameters ($w_\text{D}$, $w_\text{C}$, $w_\text{E}$, $w_\text{p}$) $=$ (1, 20, 0, 0.01).
In the first pass, and for optical flow only, we speed up computation using precomputed \textsc{daisy} descriptors for all pixels.

%%--------------------------------------------------------------------------------------------------
\subsection{Laplacian occlusion filling}
\label{sec:LaplacianFilling}

\begin{figure}
	\input{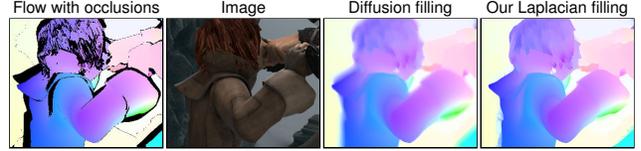}
	\caption{\label{fig:fill-in-example}%
		Our Laplacian occlusion filling preserves images edges in the filled flow fields better than a simple diffusion fill.
		Data from MPI-Sintel \cite{ButleWSB2012}.
	}\vspace{-0.5em}
\end{figure}

%% Motivation.
Pairwise correspondences are incorrect in areas of occlusion, where a point is only visible in one view, but not the other.
In these cases, the correspondences computed in the previous section are mismatches, which would negatively impact the variational scene flow estimation in the next section.
We therefore invalidate and fill in occluded pixels in the stereo flows.
As wide-baseline views can cause large occlusion regions, simple occlusion filling strategies, such as diffusion or weighted median filtering, cannot handle them adequately.
We propose a new occlusion filling method based on the observation that flow values are linearly correlated with colour intensities within a small window, which we exploit using local linear regression.
This gives more cues for the occlusion filling than just using the surface.
\cref{fig:fill-in-example} illustrates this: most edges are sharp in our filled flow field.
However, note that the flow is imperfect on the girl's back due to the lack of strong image edges in this area.
In such cases, our filling result degrades gracefully, and still is better than diffusion.

%% Invalidation.
We begin by computing a binary occlusion mask from our bidirectional stereo flows using the forward-backward check with a threshold of 3\,pixels.
This mask, which can optionally be cleaned using morphological closing, determines which pixels are invalidated and will be filled in in the next step.

\begin{figure}[t]
	\input{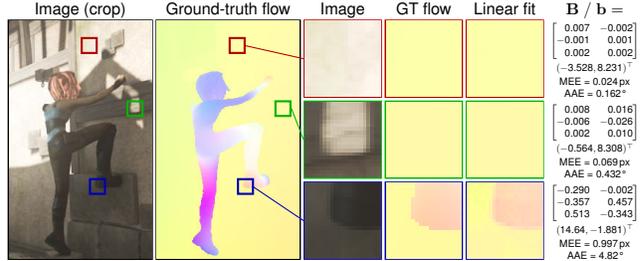}
	\caption{\label{fig:LocalLinearity}%
		Visualisation of the local linearity between 8-bit RGB colours in a 21$\times$21 window and the corresponding flow values (\cref{eq:LocallyLinear}) for three examples:
		constant colour and flow (red, top), constant flow despite textured patch (green, middle), and image edges coinciding with flow discontinuities (blue, bottom).
		Error measures are mean endpoint error (MEE) and average angular error (AAE).
		Data from MPI-Sintel \cite{ButleWSB2012}.
	}\vspace{-0.5em}
\end{figure}

%% Filling.
The key assumption is that the flow $\mathbf{u}_i \!=\! \mathbf{y}_i \!-\!\mathbf{x}_i$ between corresponding points $\mathbf{x}_i$ and $\mathbf{y}_i$ can be expressed as a linear function of the image colours within a small window $w$ (in practice 3$\times$3 pixels):
\begin{equation}
\label{eq:LocallyLinear}
\mathbf{u}_i \approx \mathbf{B} I(\mathbf{x}_i) + \mathbf{b} \text{,\quad for all $i \in w$,}
\end{equation}
where $\mathbf{B}$ is a 2$\times$3 matrix, $I(\mathbf{x}_i)$ a 3$\times$1 RGB vector and $\mathbf{b}$ a 2$\times$1 vector.
\Cref{fig:LocalLinearity} visualises the good fit of local linearity for three example windows.
In practice, we use 3$\times$3 windows which provide an even better fit than the shown examples.
This linear relationship applies to every window a pixel is in, so we sum up all overlapping windows:
\begin{equation}
\label{eq:LaplacianCost1}
E = \sum_j  \sum_{i \in w_j}  \left\|
	\mathbf{u}_i - \left(\mathbf{B}_j I(\mathbf{x}_i) + \mathbf{b}_j\right)
\right\|^2 + \epsilon \cdot \left\| \mathbf{B}_j \right\|_F^2 \!\text{,}
\end{equation}
where $w_j$ is a window of pixels around $j$, $\epsilon \!=\! 10^{-4}$, and the regularisation term on $\mathbf{B}_j$ is included for numerical stability (as for a constant image, $\mathbf{B}$ and $\mathbf{b}$ cannot be determined uniquely), and for a smoother solution (since $\left\| \mathbf{B} \right\|_F^2 \!=\! 0$ implies that $\mathbf{u}$ is constant over the window).
Levin et al. \cite{LevinLW2008} provide a closed-form solution for this sort of cost function, by eliminating their equivalent of our $\mathbf{B}$ and $\mathbf{b}$ terms, which results in a quadratic cost in the unknowns $\mathbf{U}$ alone:
\begin{equation}
\label{eq:LaplacianCost2}
E' = \mathbf{U}^\top \mathbf{L} \mathbf{U} \text{,}
\end{equation}
where $\mathbf{U}$ is an $N \!\times$2 matrix with $\mathbf{u}_i^\top$ as its $i$th row, and $\mathbf{L}$ is the so-called matting Laplacian \cite[Eq. 12]{LevinLW2008}.

\begin{figure}
	\input{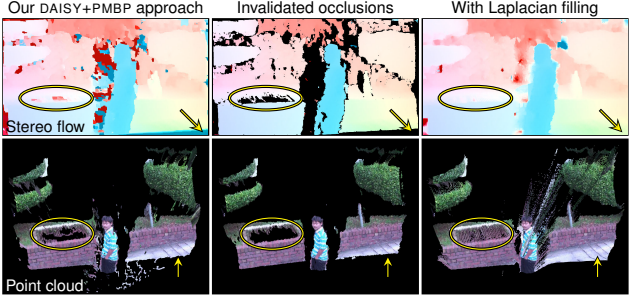}
	\caption{\label{fig:LaplacianFilling}%
		Starting from the \textsc{daisy}+\textsc{pmbp} flow (left), we invalidate occluded pixels (centre), and fill them plausibly using a novel edge-preserving Laplacian filling technique (right).
	}\vspace{-0.5em}
\end{figure}

\begin{figure}
	\scalebox{0.6}{\input{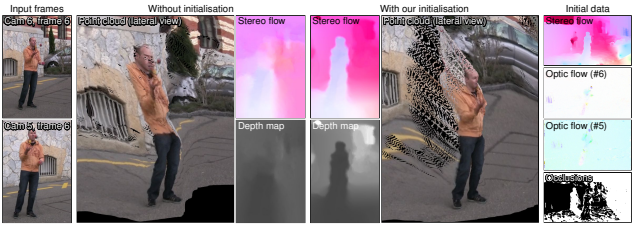}}
	\caption{\label{fig:LeviComparison}%
		Comparison of scene flow computation without (left) and with (right) initialisation from \textsc{daisy+pmbp} on the \textsc{juggler} dataset \cite{BallaBPP2010} with wide camera baseline.
		Without outside initialisation, the juggler is not reconstructed correctly as seen clearly in the point cloud, stereo flow and depth map.
		With our \textsc{daisy+pmbp} initialisation, using the flows and occlusion mask shown on the far right, the juggler is reconstructed correctly.
	}\vspace{-1em}
\end{figure}

%% Constraints.
We augment this cost with constraints and differentiate with respect to $\mathbf{U}$, resulting in the sparse linear system
\begin{equation}
\label{eq:LaplacianSystem}
(\mathbf{L} + \lambda \cdot \mathbf{D}_\text{C}) \mathbf{U} = \lambda \cdot \mathbf{U}_\text{C} \text{,}
\end{equation}
where $\mathbf{D}_\text{C}$ is a diagonal matrix with one for non-occluded pixels and zero for all other pixels, weighted by $\lambda \!=$\,5, and $\mathbf{U}_\text{C}$ is a matrix containing the specified correspondences for non-occluded pixels and zero for other pixels.
In our implementation, we compute the Cholesky factorisation of $(\mathbf{L} + \lambda \cdot \mathbf{D}_\text{C})$ once and solve \cref{eq:LaplacianSystem} separately for each column of $\mathbf{U}$.
\Cref{fig:LaplacianFilling} shows an example where areas like the ground and the grating on the left are filled with plausible flow values.

%%--------------------------------------------------------------------------------------------------
\subsection{Scene flow computation}
\label{sec:SceneFlow}

To compute the scene flow between adjacent time steps, we build upon the variational scene flow method by Valgaerts et al. \cite{ValgaBZWST2010}, which we initialise using the flows computed in the previous sections.
We take inspiration from EpicFlow \cite{RevauWHS2015}, which achieved state-of-the-art optical flow results by combining robust dense matching with a variational refinement.
This combination has two major advantages for scene flow computation.
First, when applied to wide-baseline videos with large parallax between views, the used variational approach fails to converge to the correct solution without a reasonable initialisation, which our flows provide (see \cref{fig:LeviComparison}).
Second, variational methods have the key advantage over stochastic optimisation methods like \textsc{pmbp} that their strong regularisation leads to smoother results.
In our case, the stereo flows computed with the methods in \cref{sec:DaisyPMBP,sec:LaplacianFilling} are accurate at the pixel level, but have sub-pixel noise.
When triangulating the 3D positions of these flows, the sub-pixel noise manifests itself as noisy 3D positions.

We therefore use quarter-resolution downsampled versions of our stereo and optical flows as initialisation for the variational scene flow computation, which computes smoother and refined flows at full image resolution in a multi-resolution fashion.
In short, the method of Valgaerts et al. \cite{ValgaBZWST2010} estimates the scene flow between two successive time steps by minimising an energy functional of the form
\begin{equation}
\label{eq:SceneFlowEnergy}
E = \int_\Omega \bigg(
	\underbrace{\sum_{i\!=1}^4 E_\text{D}^i}_\text{\small data} +
	\underbrace{\sum_{i\!=1}^2 \alpha_i \!\cdot\! E_\text{E}^i}_\text{\small epipolar} +
	\underbrace{\sum_{i\!=1}^3 \beta_i \!\cdot\! E_\text{S}^i}_\text{\small smoothness}
\bigg) \dif \mathbf{x} \text{.}
\end{equation}
For completeness, we discuss the individual energy terms in \ifIncludeSuppMat{\cref{sec:scene-flow}}{Supplementary \cref*{supp-sec:scene-flow}}.
We finally obtain the scene flow as the difference between the triangulated 3D positions of temporally corresponding points at time $t$ and $t\!+\!1$.
We show example results in \cref{fig:results}, with two of the four input video frames, the refined optical and stereo flows, and a visualisation of the scene flow on a point-cloud reconstruction using every hundredth scene flow vector.

%\afterpage{\clearpage}
\begin{figure*}[p]
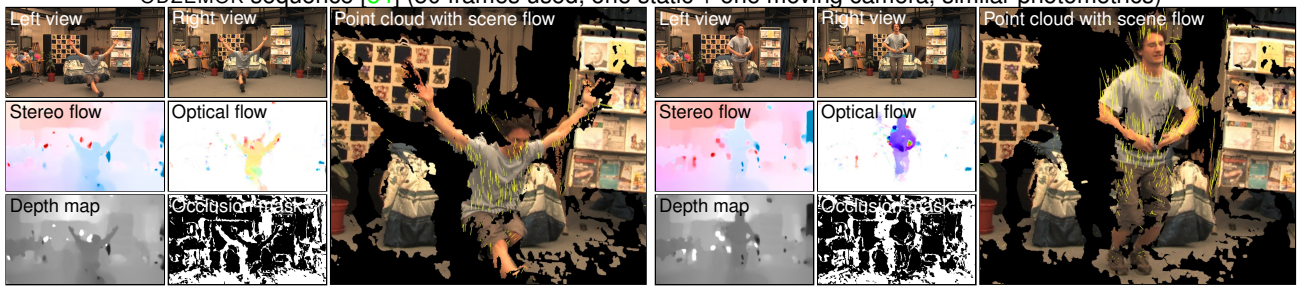

	\input{figures/deer-results-v2}\\[0.25em]
	\input{figures/boar-results}\\[0.25em]
	\input{figures/bear-results-v2}\\[0.25em]
	\input{figures/boy-results-v2}\\[0.25em]
	\input{figures/odzemok-results}
	\caption{\label{fig:results}%
		Results of our scene flow technique on the \textsc{deer}, \textsc{boar}, \textsc{bear} and \textsc{boy} datasets (all handheld), and the \textsc{odzemok} dataset.
		The smaller images show pairs of input frames (top), visualisations for stereo and optical flows (middle; scaled differently for visualisation), and depth map and occlusion mask (bottom; all for left view).
		The large images show cropped point cloud reconstructions with a subset of the scene flow vectors.
		Our approach can cope with the considerable differences in the camera and sensor characteristics in the \textsc{deer} dataset, and also the considerable camera motion and shake, which causes the changing colours of the optical flow visualisations.
	}
\end{figure*}

\begin{figure*}
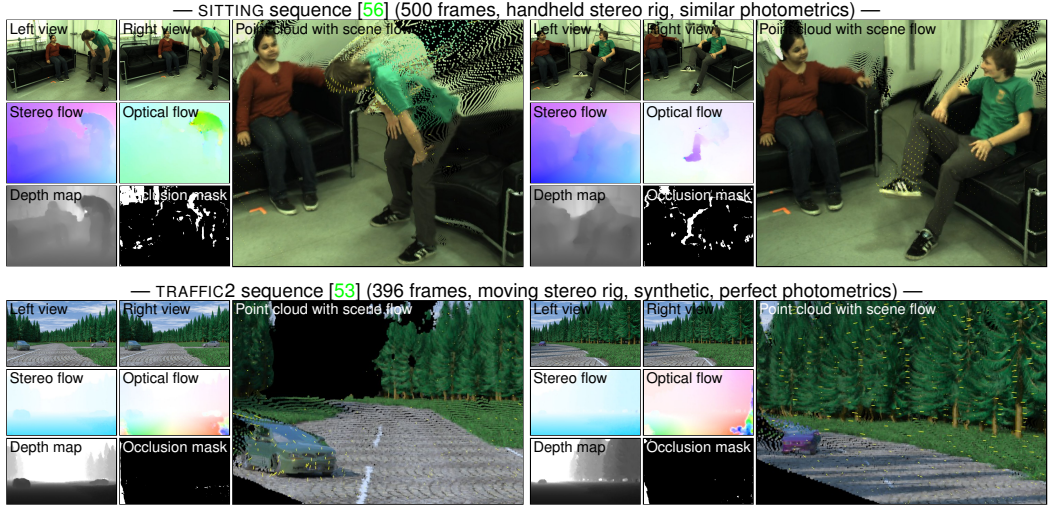
%
\centering%
\scalebox{0.8}{\input{figures/sitting-results}}\\[0.25em]
\scalebox{0.8}{\input{figures/traffic2-results}}
\caption{\label{fig:stereo-rig-results}%
	Results of our scene flow technique for moving stereo rigs on the handheld \textsc{sitting} \cite{WuSVT2013} and synthetic \textsc{traffic2} \cite{WedelBVRFC2011} datasets.
}\vspace{-1em}
\end{figure*}

%%==================================================================================================
%\clearpage
\section{Results and discussion}
\label{sec:results}

To our knowledge, we propose the first dense scene flow technique for handheld, independently moving cameras with wide baselines in both the geometric and photometric sense.
In \cref{fig:results}, we show scene flow results from handheld cameras on our \textsc{deer} and \textsc{boar} datasets, and Jiang et al.'s \textsc{bear} and \textsc{boy} datasets \cite{JiangLTZB2012}, where we show two frames covering about a second of time.
Notice the different colours and exposure in the \textsc{deer} and \textsc{boar} datasets, and how the cameras move and shake over time.
We are nonetheless able to faithfully capture dense stereo and scene flows.
The \textsc{boy} dataset, in particular, shows considerable camera motion, which can be seen in the changing colours of the optical flow visualisation.
In \cref{fig:stereo-rig-results}, we furthermore show results for moving stereo rigs, where the cameras are not moving independently, but are fixed in a rig.
\Cref{fig:JiangComparison} compares our depth map to one computed by Jiang et al.'s handheld stereo technique \cite{JiangLTZB2012}, and shows that our result has smoother depth variations thanks to our variational scene flow refinement.
Please see our supplemental video for video clips of our results.
%
%For best results, we use the camera calibration data provided with each dataset.

\begin{figure}
	\input{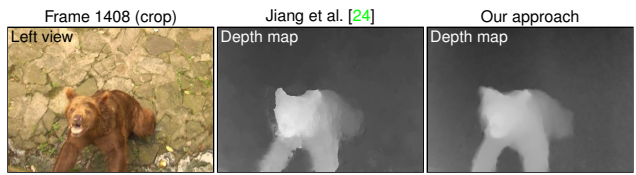}
	\caption{\label{fig:JiangComparison}%
		Comparison of depth maps for the \textsc{bear} sequence by Jiang et al. \cite{JiangLTZB2012} and our approach.
		Our variational refinement produces smoother depth variations on the bear and the background, and only slightly blurrier depth boundaries.
	}%\vspace{-1em}
\end{figure}

Most dense two-view scene flow techniques use a narrow camera baseline (tens of cm) and fail for wider baselines.
We show an example in \cref{fig:LeviComparison}, in which we compare the variational scene flow technique described in \cref{sec:SceneFlow} without and with our flow initialisation.
Without our flow initialisation, stereo and optical flows are computed in a coarse-to-fine manner over more than 50 pyramid levels.
The incremental flow refinement at each pyramid level fails to reconstruct the juggler, whereas our flow initialisation provides the necessary input to reconstruct him correctly.

\inlineheading{Occlusion filling}
We more thoroughly evaluate our Laplacian occlusion filling technique on MPI-Sintel \cite{ButleWSB2012}.
We invalidate flow pixels marked in the provided occlusion maps, fill them using our technique and compare to the ground-truth flow fields with diffusion-based filling \cite{Weick1998} as baseline.
For the first 10 frames of the 23 training datasets, Laplacian filling has a mean endpoint error of 0.92 pixels (diffusion-based: 1.34), and an average angular error of 5.7\textdegree\  (6.4\textdegree).
Laplacian filling has the smallest mean endpoint error for all 230 frames.
This means that the assumption of correlated colour and flow \textit{edges} is valid in most cases.

\begin{figure}[t]
	\centering\scalebox{0.8}{% !TEX root = ../WideBaselineSceneFlow-3DV2016.tex
\centering%
%\tikzset{external/remake next}%
\tikzsetnextfilename{gt-eval-EISATS-seq2-small}%
\begin{tikzpicture}

\begin{axis}[name=MAEd,
	footnotesize, scale only axis, width=0.52\linewidth, height=2.7cm,
	xmin=0, xmax=400, xtick={0,50,...,400}, xlabel={Frame number},
	ymin=0, ymax=2.3, ytick={0,0.5,...,2.5}, %ylabel={Euclidean distance [cm]},
	title={MAE of stereo flow: $d \!=\! \|\mathbf{u}_2\|$ [px]},
	title style={yshift=-0.5ex},
	legend columns=1,
	legend entries={\textsc{daisy+pmbp}, Laplacian filling, Variational refinement},
	legend style={font=\scriptsize,cells={anchor=west}},
]%

\addplot[red!70!black,    mark=none, semithick] table[x=Frame, y=MAE_d]  {figures/data/EISATS-eval/stats-seq2-dpm.txt};
\addplot[blue!70!black,   mark=none, semithick] table[x=Frame, y=MAE_d]  {figures/data/EISATS-eval/stats-seq2-filled.txt};
\addplot[green!60!black,  mark=none, semithick] table[x=Frame, y=MAE_d]  {figures/data/EISATS-eval/stats-seq2-lsf.txt};

\end{axis}

\begin{axis}[name=RMSE,
	at={($(MAEd.east)+(8mm,0)$)}, anchor=west,
	footnotesize, scale only axis, width=0.52\linewidth, height=2.7cm,
	xmin=0, xmax=400, xtick={0,50,...,400}, xlabel={Frame number},
	ymin=0, ymax=23, ytick={0,5,...,25}, %ylabel={Error},
	title={RMSE and AAE},
	title style={yshift=-0.2ex},
	legend columns=1, legend entries={RMSE [px], AAE [°\,]},
	legend style={font=\scriptsize,cells={anchor=west}},
]%

\addplot[red!70!black,  mark=none, semithick] table[x=Frame, y=RMSE] {figures/data/EISATS-eval/stats-seq2-lsf.txt};
\addplot[blue!70!black, mark=none, semithick] table[x=Frame, y=AAE]  {figures/data/EISATS-eval/stats-seq2-lsf.txt};

\end{axis}

\end{tikzpicture}%
}
	\caption{\label{fig:GTEvaluation}%
		Per-frame error statistics for the \textsc{traffic2} dataset \cite{WedelBVRFC2011}.
		Each step in our pipeline reduces the mean absolute error (MAE) of the estimated stereo flows (left).
		The right graph shows the total error of our scene flow, in terms of RMSE and AAE.
	}\vspace{-1em}
\end{figure}
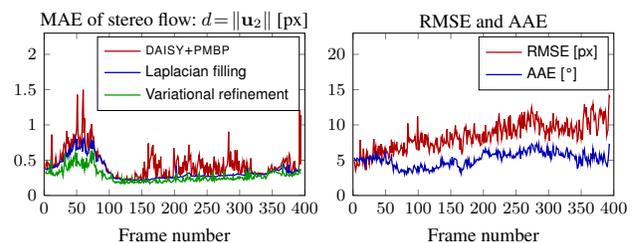

\inlineheading{Ground-truth evaluation}
There are no existing ground-truth datasets for evaluating scene flow estimated from handheld cameras, let alone the wide-baseline case our approach addresses.
We therefore quantitatively evaluate our scene flow results using the synthetic dataset \textsc{traffic2} \cite{WedelBVRFC2011} with ground-truth flows.
This dataset models a car-mounted stereo camera as the car drives along a road.
As such, the cameras are fixed in a static rig and do not move independently, but this is the closest available synthetic dataset to our application scenario.
In \cref{fig:GTEvaluation}, we plot per-frame error statistics across the whole sequence.
Using the same notation as Wedel et~al. \cite{WedelBVRFC2011}, we evaluate the stereo flow using the mean absolute error $\text{MAE}_d = \frac{1}{|\Omega|} \sum_\Omega | d - \tilde{d} |$, where $\Omega$ is the domain of all image pixels and $d \!=\! \|\mathbf{u}_2\|_2$ is the estimated\footnote{Our approach does not assume rectified stereo images and the stereo flow $\mathbf{u}_2$ is therefore not restricted to be horizontal, so we use its $\ell_2$ norm.} and $\tilde{d}$ the ground-truth disparity.
%
%This measure easily extends to the left optical flow components $(u,v) \!=\! \mathbf{u}_1$ as well as the \emph{disparity flow} $p \!=\! \|\mathbf{u}_3\|_2$.
%
This metric excludes occluded regions.
The top of \cref{fig:GTEvaluation} shows that each of our processing steps reduces the error in the estimated stereo flow fields.
The scene flow is evaluated using the root mean squared error (RMSE) and the average angular error (AAE):
{\small\begin{align}
\label{eq:RMSE}
\text{RMSE} &= \sqrt{ \frac{1}{|\Omega|} \sum_\Omega \left\| (u, v, d, p) - (\tilde{u}, \tilde{v}, \tilde{d}, \tilde{p}) \right\|^2 } \text{,} \\
\text{AAE}  &= \frac{1}{|\Omega|} \sum_\Omega \cos^{-1} \frac{
	(u, v, p, 1) \cdot (\tilde{u}, \tilde{v}, \tilde{p}, 1)
}{
	\|u, v, p, 1\| \cdot \|\tilde{u}, \tilde{v}, \tilde{p}, 1\|
} \text{.}
\end{align}\par}

\inlineheading{Artificial contamination}
To demonstrate the robustness of our method to photometric differences, we contaminated one video of the \textsc{bear} dataset, and still obtained comparable reconstruction results to the clean input videos (\cref{fig:bear-contaminated}).

\begin{figure}[t]
	\input{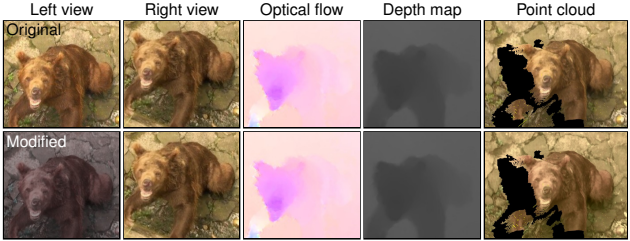}
	\caption{\label{fig:bear-contaminated}%
		Our approach is robust to photometric differences in the input videos: with modified input colours (bottom), we obtain results comparable to the clean videos (top).
		}\vspace{-1em}
\end{figure}

\inlineheading{Runtimes}
Our correspondence finding takes about six minutes for computing optical flow or stereo flow between an image pair with 960$\times$540 resolution.
The occlusion filling takes 7 seconds, and the variational refinement 1.6 minutes for each set of four frames (single-threaded run times on a 3.5\,GHz Xeon CPU).

\inlineheading{Extrinsic self-calibration}
We experimented with estimating epipolar geometry, and hence relative extrinsic calibration, in our \textsc{daisy+pmbp} approach in \cref{sec:DaisyPMBP}, but we found that a global extrinsic calibration of the input videos using structure-from-motion produces more stable results, as all cameras share the same global coordinate system.

\inlineheading{Failure cases}
Like all scene flow techniques, our proposed technique fails if the stereo or optical flows contain incorrectly matched correspondences.
This can for example happen if scene motion is too fast, leading to inaccurate optical flow estimation, which impacts the quality of the scene flow estimation.
We show such a case for two camera views from the \textsc{breakdancers} dataset \cite{ZitniKUWS2004} in \cref{fig:FailureCases}.
The main source of mismatches in our optical and stereo flows are large areas of constant colour, which are not sufficiently regularised by \textsc{pmbp}, and incorrect correspondences in occlusion regions that are not detected as such.
The resulting spurious scene flow can be seen in \cref{fig:FailureCases}, on the floor and the background.

\begin{figure}[t]
	\input{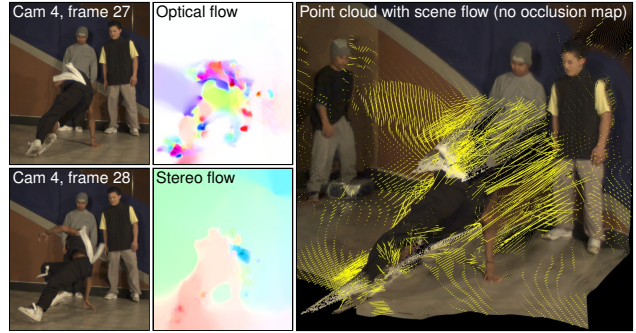}
	\caption{\label{fig:FailureCases}%
	Example of a failure case: the motion between frames of the \textsc{breakdancers} dataset \cite{ZitniKUWS2004} (using two out of eight cameras) is too fast (see images on the left), leading to poor optical flow estimation and hence poor scene flow.
	}\vspace{-1em}
\end{figure}

\inlineheading{Limitations}
Camera baselines wider than about 30° prove problematic, but could potentially be handled with affine-invariant descriptors.
We also observe that depth discontinuities are somewhat blurred spatially, which results in `rubber sheet' artefacts.
This could be addressed using edge-aware regularisation.

%%==================================================================================================
\inlineheading{Discussion}
We understand our work as a step towards unconstrained dynamic 3D scene reconstruction in general environments and from just a few handheld videos.
In practice, most mobile cameras are not calibrated and most videos are not synchronised, so an important direction for the future are algorithms that can jointly estimate calibration and synchronisation parameters, in addition to the dynamic geometry.
Perhaps one could also exploit the increased temporal resolution obtained from rolling shutter or when cameras are not frame-synchronised.
Real videos are also affected by motion blur, lens flares and changing lens parameters such as zoom and aperture, which all have a negative impact on correspondence finding, but could also be exploited to gain additional information about the scene.

%%==================================================================================================
\section{Conclusion}
\label{sec:conclusion}

%% Summary + Contributions.
We presented a dense scene flow technique for two general handheld videos, captured with wide camera baseline, and different camera and sensor characteristics.
Our technique supports wider baselines than previous dense scene flow techniques by virtue of a novel wide-baseline correspondence finding approach built on \textsc{daisy} descriptors and colour consistency adjustment with \textsc{pmbp} optimisation.
We improve stereo and optical flows computed in the process of scene flow estimation using a new edge-aware Laplacian occlusion filling method that exploits image information to complete previously invalided occluded pixels.
We finally refine all flows in a variational scene flow formulation, to obtain dense, smooth correspondences across space and time.
This combination of techniques enables dense scene flow and stereo geometry computation from handheld videos, which we demonstrated on a range of challenging datasets with complex motions.

\inlineheading{Acknowledgements}
We thank the authors of the used datasets.
Funded by ERC Starting Grant 335545 CapReal.

\ifIncludeSuppMat{\appendix% !TEX root = WideBaselineSceneFlow-supplemental-3DV2016.tex
% !TeX spellcheck = en_GB
% !TEX spellcheck = en_GB
%%==================================================================================================

\section{Variational scene flow computation}
\label{sec:scene-flow}

\begin{figure}[b]
	% !TEX root = ../WideBaselineSceneFlow-supplemental-3DV2016.tex
\centering%
%\tikzset{external/remake next}%
\tikzsetnextfilename{sceneflow-geometry}%
\begin{tikzpicture}[
%	add font={\fontsize{7}{7}},
%	show background rectangle,
	image/.style={inner sep=0pt, shape=rectangle,fill=blue!20,draw=blue!50!black,very thick,rounded corners=1mm,minimum width=26mm, minimum height=14mm}]

%% Image boxes.
\node[name=L1,image,rotate=-3] at (0,0) {};
\node[name=L2,image,rotate=-7,above right=9mm and 1mm,anchor=south] at (L1.north) {};
\node[name=R1,image,rotate=3,right=18mm] at (L1.east) {};
\node[name=R2,image,rotate=6,right=14mm] at (L2.east) {};

%% Image labels.
\node[left=1mm,  inner xsep=0mm] at (L1.west) {$I_1^t$};
\node[left=1mm,  inner xsep=0mm] at (L2.west) {$I_1^{t+1}$};
\node[right=1mm, inner xsep=0mm] at (R1.east) {$I_2^t$};
\node[right=1mm, inner xsep=0mm] at (R2.east) {$I_2^{t+1}$};

%% Epipolar lines.
\draw[shorten >=.2mm, shorten <=.2mm] (L1.190) -- (L1.340) node[pos=0.3] (PL1) {};
\draw[shorten >=.2mm, shorten <=.2mm] (L2.170) -- (L2.355) node[pos=0.35] (PL2) {};
\draw[shorten >=.2mm, shorten <=.2mm] (R1.185) -- (R1.10) node[pos=0.6] (PR1) {};
\draw[shorten >=.2mm, shorten <=.2mm] (R2.180) -- (R2.15) node[pos=0.65] (PR2) {};

%% Original point positions.
\foreach \point in {L1,L2,R1,R2}
{
	\node (X\point) at ($(\point)-(L1)+(PL1)$) {};
}

%% Points on epipolar lines (and arrows to them).
\foreach \point in {L2,R1,R2}
{
	\draw[->, thick] (X\point) -- (P\point);
	\fill[black] (P\point) circle[radius=.7mm];
}

%% Draw points on epipolar lines.
\foreach \point in {L1,L2,R1,R2}
{
	\fill[blue!70!black] (X\point) circle[radius=.7mm];
}

%% Point labels.
\node[above,rotate=-3] at (PL1) {$\mathbf{x}$};
\node[above right=0mm and -2mm,rotate=-7] at (PL2) {$\mathbf{x} \!+\! \mathbf{u}_1$};
\node[above right=0mm and -5mm,rotate=3] at (PR1) {$\mathbf{x} \!+\! \mathbf{u}_2$};
\node[above right=-1.7mm and -16.5mm,rotate=6] at (PR2) {$\mathbf{x} \!+\! \mathbf{u}_1 \!+\! \mathbf{u}_2 \!+\! \mathbf{u}_3$};

%% Arrows between frames.
\draw[<->,very thick] (L1.50)  -- (L2.310) node [midway, left=1mm, rotate=-5, align=right] {\small optical flow\\[-1.5mm]$\mathbf{u}_1$};
\draw[<->,very thick] (R1.130) -- (R2.230) node [midway, right=1mm, rotate=4.5, align=left] {\small optical flow\\[-1.5mm]$\mathbf{u}_1 \!+\! \mathbf{u}_3$};
\draw[<->,very thick] (L1.15)  -- (R1.165) node [midway, below=1mm, align=center] {\small stereo\\[-1.5mm]$\mathbf{u}_2$};
\draw[<->,very thick] (L2.345) -- (R2.195) node [midway, above=1mm, align=center] {\small stereo\\[-1.5mm]$\mathbf{u}_2 \!+\! \mathbf{u}_3$};

\end{tikzpicture}%
	\caption{Four-frame configuration used in scene flow computation.}\vspace{-1em}
	\label{fig:SceneFlowGeometry}
\end{figure}
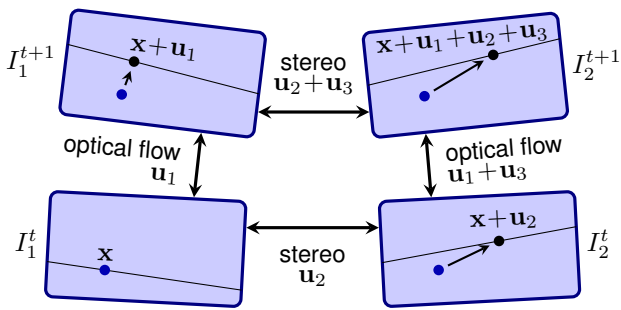

The method of Valgaerts et al. \cite{ValgaBZWST2010} estimates the scene flow between two successive time steps by minimising an energy functional of the form
\begin{equation}
\label{eq:SceneFlowEnergy}
E = \int_\Omega \bigg(
\underbrace{\sum_{i\!=1}^4 E_\text{D}^i}_\text{\small data} +
\underbrace{\sum_{i\!=1}^2 \alpha_i \!\cdot\! E_\text{E}^i}_\text{\small epipolar} +
\underbrace{\sum_{i\!=1}^3 \beta_i \!\cdot\! E_\text{S}^i}_\text{\small smoothness}
\bigg) \dif \mathbf{x} \text{.}
\end{equation}
The first part of this energy collects four data terms that measure the difference in brightness between corresponding points in the four-frame configuration of \cref{fig:SceneFlowGeometry}:
\begin{align}
\label{eq:SceneFlowDataTerm1}
E_\text{D}^1 &= \Psi\!\left(\|
I_1^{t\!+\!1}(\mathbf{x} \!+\! \mathbf{u}_1) -
I_1^t(\mathbf{x})
\|_2^2\right) \!\text{,}\\[-.2em]
\label{eq:SceneFlowDataTerm2}
E_\text{D}^2 &= \Psi\!\left(\|
I_2^{t\!+\!1}(\mathbf{x} \!+\! \mathbf{u}_1 \!+\! \mathbf{u}_2 \!+\! \mathbf{u}_3) \!-\!
I_2^t(\mathbf{x} \!+\! \mathbf{u}_2)
\|_2^2\right) \!\text{,}\\[-.2em]
\label{eq:SceneFlowDataTerm3}
E_\text{D}^3 &= \Psi\!\left(\|
I_2^t(\mathbf{x} \!+\! \mathbf{u}_2) -
I_1^t(\mathbf{x})
\|_2^2\right) \!\text{,}\\[-.2em]
\label{eq:SceneFlowDataTerm4}
E_\text{D}^4 &= \Psi\!\left(\|
I_2^{t\!+\!1}(\mathbf{x} \!+\! \mathbf{u}_1 \!+\! \mathbf{u}_2 \!+\! \mathbf{u}_3) \!-\!
I_1^{t\!+\!1}(\mathbf{x} \!+\! \mathbf{u}_1)
\|_2^2\right) \!\text{.}
\end{align}
Here, $\mathbf{u}_1$ and $\mathbf{u}_2$ denote the optical flow in the first view and the stereo flow at time $t$, respectively, while $\mathbf{u}_3$ closes the correspondence loop from $I_1^t$ to $I_2^{t\!+\!1}$.
The images $I_1^t$ are colour-corrected to match $I_2^t$ using the transform $[\mathbf{A} \ \ \mathbf{a}]$ estimated for \ifIncludeSuppMat{\cref{eq:ColourTerm}}{\cref{main-eq:ColourTerm} in the main paper} – without this appearance normalisation, matching would be much harder.
To handle the remaining appearance differences, we also include the gradient difference for improved matching in the presence of noise and lighting changes over time.
We also disable the data terms for pixels that are marked as occluded in the occlusion mask, so that their flow value is chiefly determined by the epipolar and smoothness terms.
For all terms, $\Psi(s^2) \! = \! \sqrt{s^2 \! + \! 10^{-6}}$ is the regularised $\ell_1$ penaliser.
We use $(\alpha_i, \beta_1, \beta_2, \beta_3) \!=\! (\text{10, 31, 60, 200})$ for all results.

The second term of the energy favours correspondences that satisfy the epipolar constraint between $I_1$ and $I_2$:
\begin{align}
\label{eq:SceneFlowGeometryTerm1}
E_\text{E}^1 &= \Psi\!\left(\left(
(\mathbf{x} \!+\! \mathbf{u}_2)^{\!\top} \mathbf{F}_{\!t} \, \mathbf{x}
\right)^2\right) \!\text{,}\\[-.2em]
\label{eq:SceneFlowGeometryTerm2}
E_\text{E}^2 &= \Psi\!\left(\left(
(\mathbf{x} \!+\! \mathbf{u}_1 \!+\! \mathbf{u}_2 \!+\! \mathbf{u}_3)^{\!\top} \mathbf{F}_{\!t+1} (\mathbf{x} \!+\! \mathbf{u}_1)
\right)^2\right) \!\text{,}
\end{align}
where $\mathbf{F}_t$ and $\mathbf{F}_{t+1}$ are the fundamental matrices at times $t$ and $t\!+\!1$.
Note that the variational formulation uses different data and epipolar terms than our matching cost (\ifIncludeSuppMat{\protect\cref{eq:MatchingCost}}{\protect\cref{main-eq:MatchingCost} in main paper}), as the terms used in our variational formulation are sufficient when provided with a good initialisation, as in our case.

The last term imposes regularized total-variation smoothness – the standard TV norm is defined as $\left\|\nabla \mathbf{u}\right\|_2$ \cite{RudinOF1992} – on the estimated flows by penalising their spatial derivatives:
\begin{align}
\label{eq:SceneFlowSmoothnessTermTV}
E_\text{S}^i &= \Psi\!\left(\left\|
\nabla \mathbf{u}_i
\right\|_2^2\right) \!\text{,} \quad \text{ for } i = \text{1, 2, 3.}
\end{align}

%%==================================================================================================
\section{Camera motion in the used datasets}
\label{sec:camera-motion}

Most of the datasets we use in our paper (\textsc{bear}, \textsc{boar}, \textsc{boy}, \textsc{deer}) were captured with independently moving, handheld cameras.
This is clearly visible when looking at the camera baselines and angles between cameras over time, which are shown in \cref{fig:baselines}.
The camera baselines vary by more than 50 percent, and up to 250 percent (\textsc{deer}), while the angle between cameras varies over a range of 4 degrees (\textsc{boar}) to 36 degrees (\textsc{deer}).
The \textsc{odzemok} dataset has a constant camera baseline, but the angle between cameras varies between about 10 and 20 degrees.
The \textsc{traffic2} dataset (now shown in \cref{fig:baselines}) uses a fixed stereo calibration with constant baseline and parallel cameras for all video frames.

\begin{figure}
	\centering
	\begin{tikzpicture}[tight background]
%	every axis title/.append style={at={(0.85,0.6)}},]%,show background rectangle]%, background rectangle/.style={fill=lightgray}]
	\begin{groupplot}[
	group style={group size=1 by 5, vertical sep=10mm, xlabels at=edge bottom},
	footnotesize, width=0.6\columnwidth, height=35mm,
	xlabel={Frame number},
%	xlabel shift=-4pt,
%	every axis x label/.append style={at={(ticklabel* cs:0.65)}},
%	tick label style={
%		/pgf/number format/assume math mode=true},
	ytick={1.0, 1.5, 2.0, 2.5},
	table/every table/.style={header=true, x index=0, y index=3}%,x expr=\coordindex}
	]
	
	\nextgroupplot[title={\textsc{bear} (36 frames)}, %xmin=1236, xmax=1410,
		xmin=1375, xmax=1410, xtick={1380, 1400}, ymax=1.55]
	\addplot[blue!70!black,  mark=none, semithick] table {figures/data/camera-motion/bear-camera-motion.txt};

	\nextgroupplot[title={\textsc{boar} (69 frames)}, xmin=0, xmax=68, xtick={0,25,...,60}, ymax=1.55]
	\addplot[blue!70!black,  mark=none, semithick] table {figures/data/camera-motion/boar4b-B-D-camera-motion.txt};
	
	\nextgroupplot[title={\textsc{boy} (131 frames)}, xmin=0, xmax=130, xtick={0,50,...,100}]
	\addplot[blue!70!black,  mark=none, semithick] table {figures/data/camera-motion/boy-camera-motion.txt};
	
	\nextgroupplot[title={\textsc{deer} (200 frames)}, xmin=10, xmax=209, xtick={0,50,...,200}]
	\addplot[blue!70!black,  mark=none, semithick] table {figures/data/camera-motion/deer4a-B-C-camera-motion.txt};
	
	\nextgroupplot[title={\textsc{odzemok} (50 frames)}, xmin=183, xmax=232, xtick={190,200,...,230}, ymax=1.55]
	\addplot[blue!70!black,  mark=none, semithick] table {figures/data/camera-motion/odzemok-2-6-camera-motion.txt};
	
%	\nextgroupplot[title={\textsc{sitting} (500 frames)}, xmin=0, xmax=499, xtick={0,100,...,400}, ymax=1.55]
%	\addplot[blue!70!black,  mark=none, semithick] table {figures/data/camera-motion/chenglei_rajvi_helge_sit-camera-motion.txt};
	
	\end{groupplot}
	\end{tikzpicture}%
	\begin{tikzpicture}[tight background]
		\begin{groupplot}[
			group style={group size=1 by 5, vertical sep=10mm, xlabels at=edge bottom},
			footnotesize, width=0.6\columnwidth, height=35mm,
			xlabel={Frame number},
			table/every table/.style={header=true, x index=0, y index=2}
		]
		
			\nextgroupplot[title={\textsc{bear} (36 frames)}, %xmin=1236, xmax=1410,
				xmin=1375, xmax=1410, xtick={1380, 1400}]
			\addplot[blue!70!black,  mark=none, semithick] table {figures/data/camera-motion/bear-camera-motion.txt};
			
			\nextgroupplot[title={\textsc{boar} (69 frames)}, xmin=0, xmax=68, xtick={0,25,...,60}]
			\addplot[blue!70!black,  mark=none, semithick] table {figures/data/camera-motion/boar4b-B-D-camera-motion.txt};
			
			\nextgroupplot[title={\textsc{boy} (131 frames)}, xmin=0, xmax=130, xtick={0,50,...,100}]
			\addplot[blue!70!black,  mark=none, semithick] table {figures/data/camera-motion/boy-camera-motion.txt};
			
			\nextgroupplot[title={\textsc{deer} (200 frames)}, xmin=10, xmax=209, xtick={0,50,...,200}]
			\addplot[blue!70!black,  mark=none, semithick] table {figures/data/camera-motion/deer4a-B-C-camera-motion.txt};
			
			\nextgroupplot[title={\textsc{odzemok} (50 frames)}, xmin=183, xmax=232, xtick={190,200,...,230}]
			\addplot[blue!70!black,  mark=none, semithick] table {figures/data/camera-motion/odzemok-2-6-camera-motion.txt};
	
		\end{groupplot}
	\end{tikzpicture}%
	\caption{\label{fig:baselines}%
		Visualisation of independent camera motion.
		\textbf{Left:} The baseline between cameras over time, normalised so that the minimum baseline is equal to one.
		\textbf{Right:} Angle between cameras over time (in degrees), specifically the angle between the principal axes of both cameras.
	}
\end{figure}
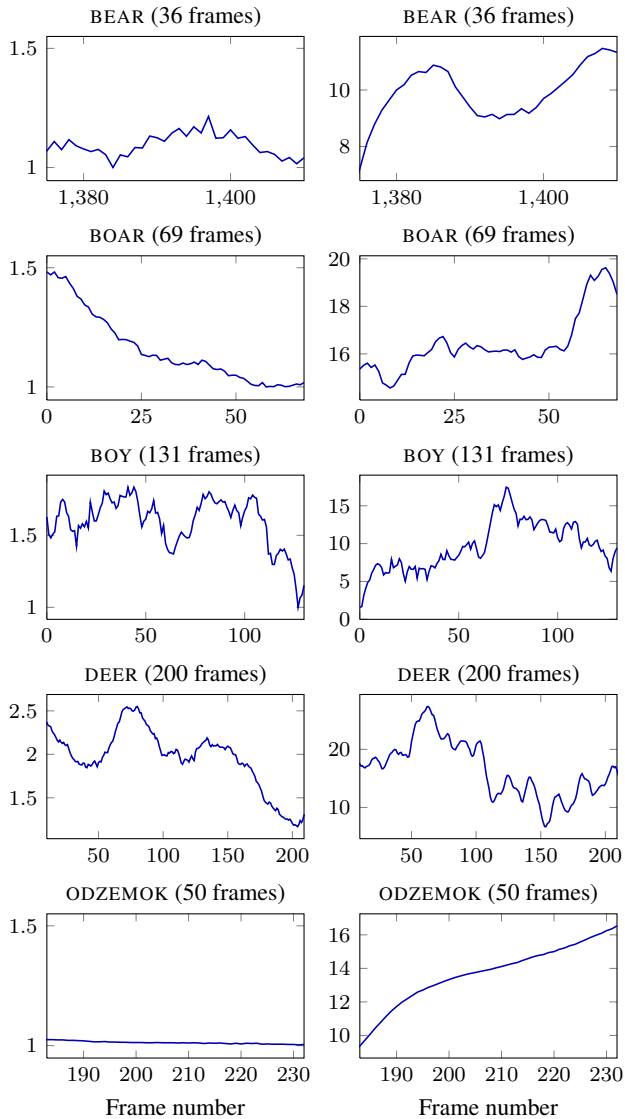
}{}

{\small
	\bibliographystyle{ieeenat}
	\bibliography{WideBaselineSceneFlow}
}

\end{document}